\documentclass[letterpaper]{article} 
\usepackage{aaai22}  
\usepackage{times}  
\usepackage{helvet}  
\usepackage{courier}  
\usepackage[hyphens]{url}  
\usepackage{graphicx} 
\usepackage{natbib}  
\usepackage{caption} 
\usepackage{algorithm}
\usepackage{algorithmic}

\usepackage{array}
\usepackage{stfloats}

\usepackage{booktabs}
\usepackage{multicol}
\usepackage{multirow}
\usepackage{makecell}

\usepackage{amsfonts}
\usepackage{mathrsfs}

\usepackage{amsmath}

\usepackage{color}

\usepackage{newfloat}
\usepackage{listings}
\DeclareCaptionStyle{ruled}{labelfont=normalfont,labelsep=colon,strut=off} 
\floatstyle{ruled}
\newfloat{listing}{tb}{lst}{}
\floatname{listing}{Listing}
\title{Fed-TDA: Federated Tabular Data Augmentation on Non-IID Data}

\author{
    Shaoming Duan\textsuperscript{\rm 1}\textsuperscript{\rm 2}, Chuanyi Liu\textsuperscript{\rm 1}\textsuperscript{\rm 2}\textsuperscript{\rm 3}\thanks{The corresponding author: liuchuanyi@hit.edu.cn.}, Peiyi Han\textsuperscript{\rm 1}\textsuperscript{\rm 2}\textsuperscript{\rm 3}, Tianyu He\textsuperscript{\rm 1}\textsuperscript{\rm 2}, Yifeng Xu\textsuperscript{\rm 1}, Qiyuan Deng\textsuperscript{\rm 1}, Yuhao Zhang\textsuperscript{\rm 1}, Xinyu zha\textsuperscript{\rm 1}
}
\affiliations{

     \textsuperscript{\rm 1}School of Computer Science, Harbin Institute of Technology (Shenzhen), Shenzhen 518055, China\\
    \textsuperscript{\rm 2}Insititute of Data Security, Harbin Institute of Technology (Shenzhen), Shenzhen 518055, China\\
    \textsuperscript{\rm 3}Department of New Networks, Peng Cheng
Laboratory, Shenzhen 518000, China\\

}

\usepackage{bibentry}

\begin{document}

\maketitle

\begin{abstract}

Non-independent and identically distributed (non-IID) data is a key challenge in federated learning (FL), which usually hampers the optimization convergence and the performance of FL. Existing data augmentation methods based on federated generative models or raw data sharing strategies still suffer from low performance, privacy protection concerns, and high communication overhead in decentralized tabular data. To tackle these challenges, we propose a federated tabular data augmentation (Fed-TDA) method that synthesizes fake tables using some low-dimensional statistics (e.g., distributions of each column and global covariance). Specifically, we propose the multimodal distribution transformation and inverse cumulative distribution mapping to synthesize continuous and discrete columns in tabular data according to the pre-learned statistics, respectively. Furthermore, we theoretically analyze that our Fed-TDA not only preserves data privacy but also maintains the distribution of the original data and the correlation between columns. Through extensive experiments on five real-world tabular datasets, we demonstrate the superiority of Fed-TDA over the state-of-the-art methods in test performance, statistical similarity, and communication efficiency. Code is available at https://github.com/smduan/Fed-TDA.git.

\end{abstract}

\section{Introduction}

In real-world applications, tabular data is the most common data type \cite{shwartz2022tabular}, which has been widely used in many relational database-based applications, such as medicine, finance, manufacturing, climate science, etc. Numerous organizations are using these data and machine learning (ML) to optimize their processes and performance. The wealth of data provides huge opportunities for ML applications. However, most of these tabular data are highly sensitive and typically distributed across different organizations. Due to privacy and regulatory concerns, these organizations are reluctant to share their private data. 

In response to these concerns, federated learning (FL) ~\cite{mcmahan2017communication,li2020federated} has been extensively studied in recent years, where multiple participants jointly train a shared deep learning model under the coordination of a central server without transmitting their local data. 
Since FL provides a feasible solution for data privacy, confidentiality,  and data ownership, it has been applied in various tabular data applications, e.g., credit card detection \cite{duan2022fed}, drug discovery \cite{chen2021fl}, and traffic flow prediction \cite{qi2021privacy}. However, a key challenge for FL is that the data of different parties are non-independent and identically distributed (non-IID), which can seriously degrade the performance of FL~\cite{li2021model,li2022federated}.

Existing studies for solving the non-IID problem can be roughly divided into two categories: \textit{algorithm-based methods} and \textit{data-based methods}. Algorithm-based methods mitigate the weight drift between clients and the server by designing appropriate loss functions \cite{li2020federatedopt,li2021model} or personalized federated learning strategies \cite{zhu2021federated,tan2022towards}. However, both our experimental results and the studies in \cite{li2022federated} show that these methods do not always perform better than vanilla FedAvg \cite{mcmahan2017communication}. Data-based methods perform data augmentation by sharing raw data \cite{yoon2020fedmix,jeong2020hiding,oh2020mix2fld} or generators \cite{wen2020unified,augenstein2019generative,wen2022communication}, which convert non-IID data to IID data and fundamentally eliminate data distribution shifts. Compared with algorithm-based methods, numerous studies on image datasets show that data-based methods can effectively improve the performance of FL \cite{yoon2020fedmix,luo2021no,wen2022communication}. 

Unfortunately, existing federated data augmentation methods directly applied to tabular data still suffer from the following challenges. Firstly, these methods fail to achieve high performance on tabular data. Unlike images, tabular data usually consists of a mixture of continuous and discrete columns. Synthesizing such mixed types of data remains a challenge for GAN-based methods, especially in terms of column distributions and correlations between columns \cite{xu2019modeling,lee2021invertible}. Furthermore, non-IID data will further degrade the performance of federated GANs \cite{zhao2021fed}. Secondly, since that tabular data usually contains sensitive information, sharing raw data or trained generators with other parties will compromise the privacy of the private data. Finally, exchanging raw data or generators between the server and clients will impose significant additional communication overhead.

To tackle these challenges, we propose Fed-TDA, a federated tabular data augmentation method for addressing the non-IID problem by synthesizing fake tables. The core idea of Fed-TDA is to synthesize tabular data with only some low-dimensional statistics (e.g., distributions of each column and global covariance). To maintain the distribution of the original data, we propose multimodal distribution transformation (MDT) and inverse cumulative distribution mapping (ICDM) to convert the random data following the standard normal distribution into continuous data following the multimodal distribution and discrete data following the cumulative distribution, respectively. In addition, to preserve the correlation between columns in the original data, our third innovation is to synthesize tabular data through a covariance-constrained data synthesis method based on global covariance, ICDM, and MDT. Unlike GAN-based methods, our Fed-TDA is a statistical model and will not be affected by non-IID data. In particular, we provide a theoretical analysis that our Fed-TDA not only protects data privacy but also preserves the distributions of the original data and the correlation between the columns.

Our main contributions can be summarized as follows:

\begin{enumerate}
\item[(1)] We propose Fed-TDA, a federated data augmentation method for solving the non-IID problem in decentralized tabular data. To the best of our knowledge, this is the first attempt to study federated data augmentation on non-IID tabular data.

\item[(2)] We propose a novel federated tabular data synthesis method for data augmentation based on some low-dimensional statistics. We theoretically provide a utility guarantee and privacy guarantee for our method.

\item[(3)] We evaluate our Fed-TDA on five real-world datasets with various data distributions among participants. The experimental results show the superiority of our Fed-TDA in terms of test performance, statistical similarity, and communication efficiency.
\end{enumerate}

\section{Related Work and Background}

\subsection{Federated Learning with non-IID Data }

In this paper, we focus on addressing the non-IID problem on decentralized tabular data. Existing relevant works can be roughly divided into two categories: algorithm-based methods and data-based methods.

\subsubsection{Algorithm-based Methods.} 
To mitigate the weight drift caused by the data distribution skew among clients, some robust loss functions are designed for FL. FedProx \cite{li2020federatedopt} adds a regularization term to the loss function to reduce the difference between the global and local models. SCAFFOLD \cite{karimireddy2020scaffold} performs control variates to correct for the weight drift in its local updates. MOON \cite{li2021model} utilizes contrastive loss to correct the training of local models. Unlike loss function-based methods, some personalized federated learning strategies are proposed to train personalized models for individual participants rather than a shared global model. In personalized federated learning, local fine-tuning \cite{tan2022towards}, meta-learning \cite{fallah2020personalized,yue2022efficient}, multi-task learning \cite{marfoq2021federated,sattler2020clustered}, and client clustering methods\cite{sattler2020clustered,ghosh2020efficient} strategies are used to learn personalized models. However, as suggested in \cite{li2022federated}, algorithm-based methods do not always perform better than vanilla FedAvg \cite{mcmahan2017communication}.

\subsubsection{Data-based Methods}

The key motivation of data-based methods is to convert non-IID data into IID by data augmentation strategies. Existing data augmentation methods can be divided into two categories: data sharing and data synthesis. Data sharing-based methods achieve data augmentation by directly sharing raw data \cite{jeong2020hiding,jeong2018communication,yoshida2020hybrid} or masked data \cite{yoon2020fedmix,oh2020mix2fld} which generated by mixup \cite{zhang2018mixup} or XOR operation. Unlike data sharing methods, data synthesis methods synthesize fake data to eliminate distribution skew by sharing a trained generative model \cite{wen2020unified,augenstein2019generative,wen2022communication}. Numerous studies have shown that data-based methods can effectively alleviate the non-IID problem on image data \cite{wen2022communication}. However, these methods still suffer from privacy leakage and high communication overhead. In particular, data-based methods fail to achieve high performance on tabular data.

\subsection{Differential Privacy}


\textbf{Definition 3.1.}\textit{(Differential Privacy \cite{dwork2014algorithmic}) A randomized algorithm $\mathcal{A}$ is $(\epsilon, \delta)$-differential privacy if for any output subset $S$ and two neighboring datasets $D$ and $D^{'}$ satisfies:}
\begin{equation} 
P(\mathcal{A}(D) \in S) \le e^{\epsilon} \cdot P(\mathcal{A}(D^{'}) \in S) + \delta
\label{equ:dp}
\end{equation}
where $\mathcal{A}(D)$ and $\mathcal{A}(D^{'})$ are the output of the algorithm for neighboring datasets $D$ and $D^{'}$, respectively, and $P$ is the randomness of the noise in the algorithm. The privacy budget $\epsilon$ is used to control the level of privacy. The smaller the $\epsilon$, the better the privacy.

The Gaussian mechanism is a classical perturbation mechanism for ($\epsilon, \delta$)-differential privacy \cite{dwork2014algorithmic}. For any deterministic vector-valued computation function $f$: $X \to R^{d}$, a Gaussian perturbation mechanism $\mathcal{M}$ is obtained by computing the function $f$ on the input data $x$ and then adds random noise sampled from a Gaussian distribution to the output $\mathcal{M}(x) = f(x)+N(0,\sigma^{2}I)$,
where $N(0,\sigma^{2}I)$ denotes a Gaussian distribution with a mean of 0 and a standard deviation of $\sigma$. For any function $f$ have the global $l_{2}$-sensitivity $\bigtriangleup = \mathop{max}\limits_{x,x^{'}}\parallel f(x)-f(x^{'})\parallel_{2}$, the Gaussian mechanism $\mathcal{M}$  with $\sigma \geq \frac{\bigtriangleup \cdot \sqrt{2ln(1.25/\delta)}}{\epsilon}$
is ($\epsilon, \delta$)-differential privacy \cite{dwork2014algorithmic}.

\section{Fed-TDA}

\subsection{Problem Formulation}


In this paper, we follow the typical assumption of FL that all participants including the server are honest-but-curious. Suppose there are $K$ clients in FL, and each client possesses a local tabular dataset $X_{k}=(x_{k},y_{k}), k=(1,2,..., K)$, where $x_{k}$ and $y_{k}$ represent the features and labels of client $k$. The goal of FL is to learn a shared model $\theta$ over the decentralized dataset $X = \bigcup_{k \in [K]}X_{k}$. The objective of vanilla FedAvg \cite{mcmahan2017communication} is:
\begin{equation} 
\mathop{\arg\min}_{\theta} \mathcal{L}(\theta) =  \sum_{k=1}^{K} \frac{N_{k}}{N} \mathbb{E}_{(x,y) \sim X_{k}}[l_{k}(\theta,(x,y))]
\label{equ:loss}
\end{equation}
where $\mathbb{E}_{(x,y) \sim X_{k}}[l_{k}(\theta,(x,y))]$ represents the empirical loss of client $k$, $N_{k}$ denotes the number of samples in client $k$, $N=\sum_{k=1}^{K}N_{k}$ is the total number of samples over all clients. From a statistical perspective, the data of $k$-th client can be denoted as $P_{x}(x_{k},y_{k})=P_{k}(x_{k}|y_{k})P_{k}(y_{k})$, which represents the joint probability between $x_{k}$ and $y_{k}$. Then, the joint distribution of global data is $P(x,y)$. In FL, we wish that the data distribution of each client is similar to the global data ($P_{i}(x_{i},y_{i})=P_{j}(x_{j},y_{j})=P(x,y), i \neq j$). 

However, in real-world applications, the data is non-IID, $P_{i}(x_{i},y_{i}) \neq P_{j}(x_{j}, y_{j}), i \neq j$, which will lead to weight drift and degrade the performance of FL. To solve this problem, we are committed to create an IID data ($P_{i}(x_{i},y_{i}) \approx P_{j}(x_{j},y_{j}) \approx P(x,y), i \neq j$) by data augmentation. The design goals of our method are as follows: (1) \textbf{High performance.} The data augmentation method needs to ensure the utility of the augmented tables in terms of data distributions and correlation between columns and further improve the performance of the FL; (2) \textbf{Privacy preserving.} The privacy of the raw data must be protected during the data augmentation processes. Specifically,  the local training data cannot be shared with other clients or the server and the augmented data cannot reveal private information; (3) \textbf{Low communication costs.} The communication resources of FL are limited, and high additional communication overhead should not be introduced in the process of data augmentation; (4) \textbf{High flexibility and scalability.} Data augmentation should be independent of the FL model so that various FL algorithms can be flexibly adapted without any modification.

\subsection{Federated Tabular Data Augmentation}

\begin{figure}[!t] 
\centering
\includegraphics[width=\linewidth]{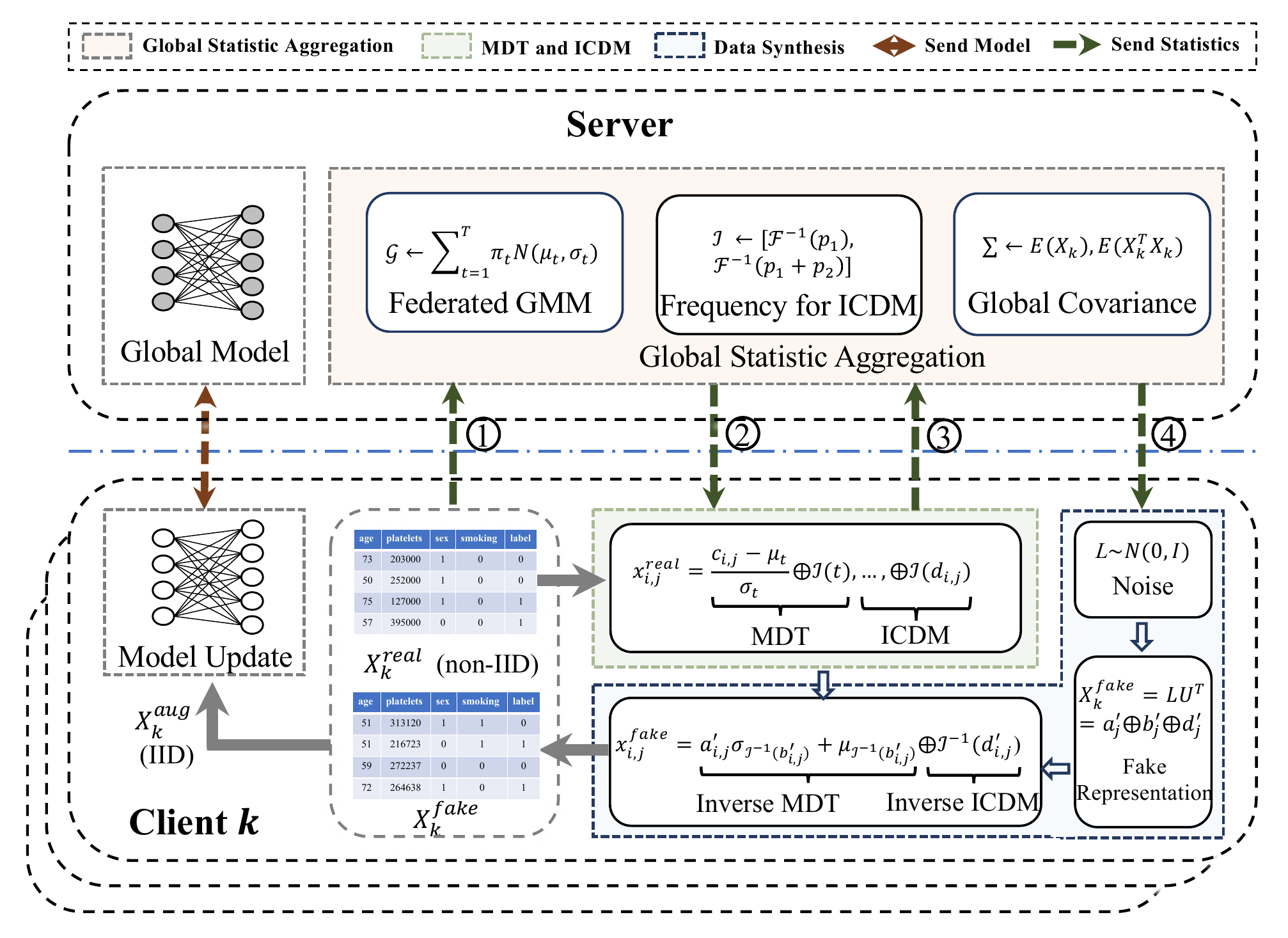}
\caption{An overview of Fed-TDA. \textcircled{1} sends local statistics, including the parameters of federated GMM and the frequency of categories. \textcircled{2} returns the parameters of federated GMM and ICDM. \textcircled{3} sends local expectations $E(X_{k})$ and  $E(X_{k}^{T}X_{k})$. \textcircled{4} returns the global covariance $\Sigma$.}
\label{fig:fedtda} 
\end{figure}


Figure \ref{fig:fedtda} shows an overview of Fed-TDA. Firstly, we compute the global multimodal distributions in continuous columns and the global frequency of categories in discrete columns from the decentralized tabular data. Secondly, we transform the tabular data into representations that follow the standard normal distribution by the proposed MDT and ICDM  locally. After that, we compute the global covariance from the transformed data representations under the coordination of the server. Finally, each client synthesizes the fake data for data augmentation from a noise matrix following the standard multivariate normal distribution based on the pre-learned statistics, MDT, and ICDM. The detailed procedures of Fed-TDA are shown in Algorithm \ref{alg:fedtdas} and \ref{alg:local}.


\begin{algorithm}[!t]
\caption{Fed-TDA (Server Side)}
\label{alg:fedtdas}
\textbf{Input}: Global data set $X=(X_{1}, X_{2},...,X_{K})$, number of clients $K$, number of federated training rounds $R$.

\textbf{Output}: Well trained model $\theta$. 

\begin{algorithmic}[1] 


\STATE $\mathcal{G} \gets$ estimate the parameters ($\mu, \sigma, \pi$) of Global GMM from each continuous column by the customized federated GMM

\STATE Compute the global frequency of categories in each discrete columns and construct the ICDM $\mathcal{I}$

\FOR{$k=0,1,...,K-1$}

\STATE Encode the local data by DMT and ICDM

\STATE Compute local expectation $E(X_{k}), E(X_{k}^{T}X_{k}) $

\ENDFOR

\STATE Compute global covariance $\Sigma$ by Equation (\ref{equ:sigma}).

\STATE $\Sigma = \Sigma + N(0,\sigma^{2} I)$

\STATE Compute the Cholesky decomposition $\Sigma = UU^{T}$

\FOR{each client $k=1,...,K$ in parallel}

\STATE $X_{k}^{'} \gets$ LocalDataSynthesis$(\mathcal{G},\mathcal{I},U)$ 

\ENDFOR

\FOR{each round $ r= 1,...,R$}
\FOR{each client $k=1,...,K$ in parallel}

\STATE $\theta_{k}^{r+1} \gets$ update local model $\theta^{r}$ using  ($X_{k}^{'} \cup X_{k}$)
\ENDFOR
\STATE $\theta^{r+1} \gets \sum_{k=1}^{K}\frac{N_{k}}{N}\theta_{k}^{r+1}$
\ENDFOR

\RETURN $\theta^{r+1}$

\end{algorithmic}
\end{algorithm}

\subsubsection{Multimodal Distribution Transformation} 

Theoretically, continuous columns generally follow a multimodal distribution \cite{lindsay1995mixture}, so any continuous column can be approximated by a finite number of Gaussian mixture distributions. In Fed-TDA, we design a multimodal distribution transformation to estimate a Mixture Gaussian Model (GMM) for each continuous column and transform them into the representations that follow the standard normal distribution. Specifically, each value from the continuous column is represented as two scalars, one representing the normalized value and the other representing the Gaussian distribution it follows. 

Firstly, we customize the variational Bayesian GMM algorithm \cite{corduneanu2001variational} in FL to learn the global GMM for each continuous column $C_{i}$. The details of the designed federated VB-GMM algorithm are shown in Appendix {\color{red}A}. From $C_{i}$, we learn a global GMM with $T$ Gaussian distributions $P_{C_{i}}(c_{i,j})=\sum_{t=1}^{T}\pi_{t}N(c_{i,j}|\mu_{t},\sigma_{t})$, where $\pi_{t}$ represents the mixture weight of \textit{t}-th Gaussian distribution. $N(c_{i,j}|\mu_{t},\sigma_{t})$ represents the \textit{t}-th Guassian distribution with mean $\mu_{t}$ and standard deviation $\sigma_{t}$. Then, each value $c_{i,j}$ in $C_{i}$ is represented as $a_{i,j} \oplus b_{i,j}$, where $a_{i,j} = \frac{c_{i,j-\mu_{t}}}{\sigma_{t}}$, $b_{i,j}=t$, and $\oplus$ represents the concatenate operation. Then, each row is represented as $x_{j}=\textit{a}_{1,j} \oplus b_{1,j} \oplus ... \oplus a_{n_{c},j} \oplus d_{1,j} \oplus ... \oplus d_{n_{d},j}$,
where $d_{i,j}$ represents the value of discrete column, $n_{c}$ and $n_{d}$ are the number of continuous and discrete columns, respectively. Note that, $b_{i,j}$ and $d_{i,j}$ represent discrete values.
 
\subsubsection{Inverse Cumulative Distribution Mapping}

\begin{figure}[!t] 
\centering
\includegraphics[width=\linewidth]{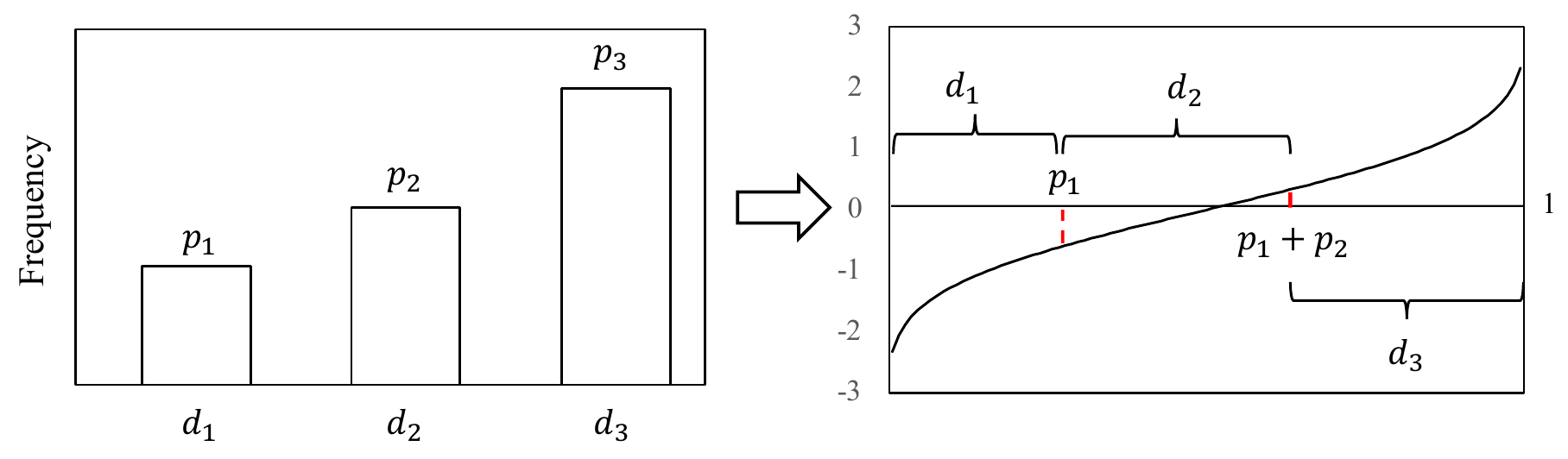} 
\caption{An example of ICDM. Categories are sorted by their frequency and then represented by a value randomly sampled from the corresponding inverse cumulative distribution of standard normal distribution.} 
\label{fig:category} 
\end{figure}

\begin{table}[!t]
\centering 
\footnotesize

\renewcommand\arraystretch{1.2}{
\setlength{\tabcolsep}{1mm}{
\begin{tabular}{|c|c|c|}
\hline

Categorical & \makecell[c]{Cumulative \\ Distribution} & \makecell[c]{Inverse Cumulative \\ Distribution} \\
\hline
$d_{1}$ & $[0+\lambda,p_{1}]$ & $[\mathcal{F}^{-1}(0+\lambda), \mathcal{F}^{-1}(p_{1})]$\\
$d_{2}$ & $[p_{1}, p_{1}+p_{2}]$ & $[\mathcal{F}^{-1}(p_{1}), \mathcal{F}^{-1}(p_{1}+p_{2})]$\\
$d_{3}$ & $[p_{1}+p_{2}, 1-\lambda]$ & $[ \mathcal{F}^{-1}(p_{1}+p_{2}), \mathcal{F}^{-1}(1-\lambda)]$ \\

\hline
\end{tabular}}}

\caption{Mapping of categories in Figure \ref{fig:category}.}
\label{tab:icdm} 
\end{table}

To model the discrete columns, we design the inverse cumulative distribution mapping to estimate the cumulative distribution for each discrete column and convert each category in the discrete columns into a continuous value that follows the standard normal distribution. 
Figure \ref{fig:category} shows an example of ICDM. Firstly, for each discrete column $D_{i}$, we compute the global frequency of each category $d_{i,j}$ from each client. For instance, in Figure \ref{fig:category}, there are three categories, namely $d_{1}$, $d_{2}$ and $d_{3}$. The global frequencies of these three categories are $p_{1}$, $p_{2}$ and $p_{3}$, respectively. Then, each category is sorted by its frequency, such as $p_{1} \leq p_{2} \leq p_{3}$ in Figure \ref{fig:category}. Finally, each category $d_{j}$ in $D_{i}$ is represented by a value randomly sampled from the corresponding interval of the inverse cumulative distribution of the standard normal distribution. Table \ref{tab:icdm} shows the inverse cumulative distribution mapping of example categories in Figure \ref{fig:category}. As shown in Table \ref{tab:icdm}, $d_{1}$ is represented as $ f_{1}$, where $f_{1} \in [\mathcal{F}^{-1}(0+\lambda), \mathcal{F}^{-1}(p_{1})]$, $\mathcal{F}^{-1}(.)$ represents the inverse cumulative distribution function of the standard normal distribution and $\lambda$ is a small parameter mainly used to prevent the output of the $\mathcal{F}^{-1}$ function from being $- \infty$ or $+ \infty$. In this paper, we set the parameter $\lambda = 10^{-4}$.

\subsubsection{Global Covariance} 

In addition to the distributions of each column, the synthetic data also needs to maintain the correlation between the columns in the original data. Therefore, we compute the global covariance $\Sigma$ from the new representations. Let $N_{k}=|X_{k}|$ represent the number of samples in client $k$, the expectation of $k$-th client is $E(X_{k}) = \frac{1}{N_{k}}\sum_{j=1}^{N_{k}}x_{k,j} \in \mathcal{R}^{1 \times l}$,
where $l$ represents the dimension of features of the normalized representation and $l=2n_{c}+n_{d}$. Then, each client send the local expectation $E(X_{k})$ to the server and compute the global expectation $E(X) = \sum_{k=1}^{K}\frac{N_{k}}{N}E(X_{k}) \in \mathcal{R}^{1 \times l}$,
where $K$ is the number of clients. According to the formula of covariance that $\Sigma_{X}=E[(X-E(X)^T(X-E(X))]$, the global covariance is 
\begin{equation} 
\Sigma_{X} = \sum_{k=1}^{K}E(X_{k}^{T}X_{k}) -(\sum_{k=1}^{K}\frac{N_{k}}{N}E(X_{k}))^{T}(\sum_{k=1}^{K}\frac{N_{k}}{N}E(X_{k}))
\label{equ:sigma}
\end{equation}
where $X_{k}^{T}X_{k}=\sum_{j=1}^{N_{k}}x_{k,j}^{T}x_{k,j} \in \mathcal{R}^{l \times l}$. 
The detailed derivation of global expectation $E(X)$ and the global covariance $\Sigma_{X}$ are provided in Appendix {\color{red}B}.

\subsubsection{Covariance-constrained Data Synthesis}


\begin{algorithm}[!t]
\caption{LocalDataSynthesis($\mathcal{G},\mathcal{I},U$)}
\label{alg:local}
\textbf{Input}: The parameters of global GMM $\mathcal{G}$, the parameter of ICDM $\mathcal{I}$, the unitary matrix $U$. 

\textbf{Output}: Synthetic data $X^{s}$. 

\begin{algorithmic}[1] 

\STATE Randomly sample $L \sim N(0,I) \in \mathcal{R}^{N_{s} \times l}$

\STATE $X^{'} = LU^{T} =[x_{i}^{'}]_{i=1}^{N_{s}} \in \mathcal{R}^{N_{s} \times l}$, where $x_{i}^{'}=a_{i,1}^{'} \oplus b_{i,1}^{'} \oplus ... \oplus a_{i,n_{c}}^{'} \oplus b_{i,n_{c}}^{'} \oplus d_{i,1}^{'} \oplus ... \oplus d_{i,n_{d}}^{'}$

\FOR{each discrete representation $b_{j}^{'}$ and $d_{j}^{'}$}

\STATE $b_{j}^{s} \gets \mathcal{I}^{-1}(b_{j}^{'})$

\STATE $d_{j}^{s} \gets \mathcal{I}^{-1}(d_{j}^{'})$

\ENDFOR

\FOR{each continuous representation $a_{j}^{'}$}

\STATE $t \gets b_{j}^{s}$

\STATE Get the parameters $\sigma_{j,t}, \mu_{j,t}$ of $t$-th Gaussian distribution from $\mathcal{G}$ 

\STATE $a_{j}^{s} \gets a_{j}^{'}\sigma_{j,t}+\mu_{j,t}$
\ENDFOR

\STATE $X^{s} = [x_{i}^{s}]_{i=1}^{N_{s}} \in \mathcal{R}^{N_{s} \times (n_{c}+n_{d})}$, where $x_{i}^{s}=a_{i,1}^{s} \oplus... \oplus a_{i,n_{c}}^{s} \oplus d_{i,1}^{s} \oplus ... \oplus d_{i,n_{d}}^{s}$

\RETURN $X^{s}$

\end{algorithmic}
\end{algorithm}


\begin{table*}[!t]
\centering
\footnotesize

  \resizebox{\textwidth}{20mm}{
\begin{tabular}{clccccccc}
\toprule
 \textbf{Datasets} &\textbf{$\beta$}&\textbf{FedAvg}& \textbf{FedProx} & \textbf{Fedmix} & \textbf{DP-FedAvg-GAN}  & \textbf{Fed-TGAN} & \makecell[l]{ \textbf{Fed-TDA} \\ \textbf{+ FedAvg}} & \makecell[l]{ \textbf{Fed-TDA} \\ \textbf{+ FedProx}} \cr
 
\midrule

 \multirow{4}{*}{Clinical} & $\beta=0.05$ & 0.605±0.032 & 0.620±0.037 & 0.635±0.043 &	0.713±0.023 & 0.897±0.038 & \textbf{0.983±0.004} & 0.947±0.018\cr
 
 & $\beta=0.1$ & 0.653±0.012 &	0.661±0.020 & 0.634±0.018	&0.831±0.018& 0.913±0.008&	\textbf{0.984±0.003} & 0.941±0.015\cr
 & $\beta=0.5$ & 0.773±0.013 &	0.764±0.025&  0.739±0.020 &	0.85±0.030 & 0.894±0.009 &	\textbf{0.986±0.003} & 0.966±0.009 \cr
 
 & $\beta=100$ (IID) & 0.975±0.007 & - & -& -& -&- &- \cr
\hline   

 \multirow{4}{*}{Tuberculosis} & $\beta=0.05$ & 0.803±0.016 &	0.806±0.031 & 0.797±0.040&0.876±0.005 &	 0.865±0.014 &0.931±0.002 & \textbf{0.939±0.004}\cr
 
 & $\beta=0.1$ & 0.867±0.011 & 0.881±0.007 & 0.877±0.007 &0.928±0.007 & 0.931±0.004 & 0.933±0.006 & \textbf{0.939±0.001}\cr
 & $\beta=0.5$ & 0.951±0.004&0.947±0.001&  0.949±0.003&0.905±0.264	& 0.956±0.002 & 0.946±0.003 & \textbf{0.961±0.000} \cr
 
 & $\beta=100$ (IID) & 0.976±0.002 & - & -& -& -&- &- \cr
 
\bottomrule
\end{tabular} }
\caption{ROCAUC scores of different methods on Clinical and Tuberculosis datasets. The parameter of Dirichlet distribution $\beta$ varies from $0.05$ to $0.5$, $K=5$, privacy budget $\epsilon= \infty$, and the local epoch is $3$.}
\label{tab:binary}
\end{table*}

\begin{table*}[!t]
\centering
\footnotesize

  \resizebox{\textwidth}{25mm}{
\begin{tabular}{clccccccc}
\toprule
 \textbf{Datasets} & \textbf{$\beta$} & \textbf{FedAvg} & \textbf{FedProx} & \textbf{Fedmix} & \textbf{DP-FedAvg-GAN} & \textbf{Fed-TGAN} & \makecell[l]{ \textbf{Fed-TDA} \\ \textbf{+ FedAvg}} & \makecell[l]{ \textbf{Fed-TDA} \\ \textbf{+ FedProx}} \cr
 
\midrule

 \multirow{4}{*}{CovType} & $\beta=0.01$ & 0.364±0 & 0.484±0.016 & 0.364±0.000 &	0.485±0.004 & 0.367±0.002& 0.529±0.019 & \textbf{0.612±0.007}\cr
 
 & $\beta=0.05$ & 0.364±0	& 0.358±0.009 & 0.607±0.003 &0.487±0.000 & 0.366±0.002 & 0.589±0.006 & \textbf{0.620±0.002}\cr
 & $\beta=0.1$ & 0.491±0.006&	0.512±0.037&  0.492±0.008& 0.570±0.005 & 0.488±0.016 & 0.600±0.003 & \textbf{0.620±0.003} \cr
 
 & $\beta=100$ (IID) & 0.649±0.008 & - & -& -& -&- &- \cr
\hline   

 \multirow{4}{*}{Intrusion} & $\beta=0.01$ & 0.590±0.125	& 0.856±0.050 & 0.575±0.074&	0.900±0.048 & 0.783±0.004	&0.956±0.008 & \textbf{0.950±0.012} \cr
 
 & $\beta=0.05$ & 0.762±0.008 & 0.970±0.005 & 0.765±0.011 &	0.967±0.009 & 0.826±0.046 &0.970±0.004 & \textbf{0.983±0.004}\cr
 & $\beta=0.1$ & 0.978±0.005	& 0.987±0.000 &  0.975±0.015 & 0.986±0.002& \textbf{0.988±0.000} & 0.980±0.002& 0.985±0.000 \cr
 
 & $\beta=100$ (IID) & 0.989±0.000 & - & -& -&- &-&- \cr
 
 \hline   

 \multirow{4}{*}{Body} & $\beta=0.01$ & 0.271±0.007 &	0.297±0.014& 0.27±0.007&0.252±0.003& 0.335±0.051  &	\textbf{0.633±0.005} & 0.605±0.016\cr
 
 & $\beta=0.05$ & 0.308±0.015&0.291±0.031 & 0.272±0.019&0.330±0.018& 0.411±0.021& \textbf{0.614±0.011} & 0.594±0.021\cr
 & $\beta=0.1$ & 0.459±0.000&0.457±0.003&  0.455±0.001	&0.459±0.007 & 0.454±0.001& \textbf{0.613±0.011} & 0.605±0.009 \cr
 
 & $\beta=100$ (IID) & 0.734±0.006& - & -& - &-&- &- \cr
 
\bottomrule
\end{tabular} }
\caption{Test accuracy of different methods on CovType, Intrusion and Body datasets. The parameter of Dirichlet distribution $\beta$ varies from $0.01$ to $0.1$, $K=5$, privacy budget $\epsilon= \infty$,  and the local epoch is $3$. }
\label{tab:multiclass}
\end{table*}

After obtaining the global covariance, each client synthesizes tabular data for data augmentation. Firstly, we inject the Gaussian noise into the global covariance $\Sigma_{X}=\Sigma_{X}+N(0,\sigma^{2}I)$ and compute its Cholesky decomposition $\Sigma_{X}=UU^{T}, U \in \mathcal{R}^{l \times l}$ on the server (lines 8-9 in Algorithm \ref{alg:fedtdas}). Then, the unitary matrix $U$ is shared to each client.

Secondly, as shown in Algorithm \ref{alg:local}, we sample a noise $L \sim N(0,I) \in \mathcal{R}^{N_{s} \times l}$ and compute the fake representation by $X^{'} = LU^{T} =[x_{i}^{'}]_{i=1}^{N_{s}} \in \mathcal{R}^{N_{s} \times l}$,  where $N_{s}$ represents the number of synthetic data. For each row $x_{i}^{'}=a_{i,1}^{'} \oplus b_{i,1}^{'} \oplus ... \oplus a_{i,n_{c}}^{'} \oplus b_{i,n_{c}}^{'} \oplus d_{i,1}^{'} \oplus ... \oplus d_{i,n_{d}}^{'}$ in $X^{'}$, we first recover the discrete columns by ICDM. For instance, for each discrete representation $b_{i,j}^{'}$ or $d_{i,j}^{'}$, if $b_{i,j}^{'}$ or $d_{i,j}^{'} \in [\mathcal{F}^{-1}(0+\lambda), \mathcal{F}^{-1}(p_{1})]$, then $b_{i,j}^{s} \gets d_{1}$ or $d_{i,j}^{s} \gets d_{1}$. Then, the continuous columns are recovered by the inverse operation of MDT, $a_{i,j}^{s}= a_{i,j}^{'}\sigma_{j,t}+\mu_{j,t}$, where $t=b_{i,j}^{s}$, $\mu_{j,t}$ and $\sigma_{j,t}$ represent the mean and the standard deviation of $t$-th Gaussian distribution, respectively. Finally, the synthetic data is represented as $X^{s}=[x_{i}^{s}]_{i=1}^{N_{s}} \in \mathcal{R}^{N_{s} \times (n_{c}+n_{d})}$, where $x_{i}^{s}=a_{i,1}^{s} \oplus... \oplus a_{i,n_{c}}^{s} \oplus d_{i,1}^{s} \oplus ... \oplus d_{i,n_{d}}^{s}$.

\subsection{Utility of Synthetic Data}
\label{sec:utility}


We ensure the utility of synthetic data from the perspective of data distribution and correlation between columns. Firstly, in Fed-TDA, the continuous and discrete columns are synthesized by MDT and ICDM, respectively. The proposed MDT and ICDM ensure that continuous and discrete columns in the synthetic data maintain the same multimodal and cumulative distributions as the original data, respectively. Secondly, Theorem 1 guarantees that the correlations between columns in the synthetic data are consistent with the original data. The detailed proof of Theorem 1 are given in Appendix {\color{red}C}.

\textbf{Theorem 1.} \textit{$X \sim N(0,I), X \in \mathcal{R}^{N \times l}$, the covariance of $X$ is $\Sigma_{X}$ and its Cholesky decomposition is $\Sigma_{X}=UU^{T}$. For $\forall L \sim N(0,I), L \in \mathcal{R}^{N \times l}$. $Y=LU^{T} \in \mathcal{R}^{N \times l}$, we have $\Sigma_{X} = \Sigma_{Y}$.}

\subsection{Privacy Analysis}

Firstly, Fed-TDA strictly follows the standard FL framework \cite{mcmahan2017communication} and protects privacy at the basic level as each client only uploads some statistics rather than the raw data. Secondly, in the data synthesis phase, the differential privacy (DP) \cite{dwork2014algorithmic} is introduced to the shared covariance. As shown in Algorithm \ref{alg:local}, the synthetic data is first constrained by global covariance. In addition, the post-processing property of DP \cite{dwork2014algorithmic} shows that any mapping of differential private output also satisfies the same-level DP. Thus, we inject Gaussian noise into the  global covariance $\Sigma = \Sigma + N(0,\sigma^{2}I)$, where $N(0,\sigma^{2}I)$ represents the Gaussian distribution with mean $0$ and standard deviation $\sigma$. To satisfy ($\epsilon,\delta$)-differential privacy, following \cite{dwork2014algorithmic}, we set $\sigma=\frac{\triangle_{\Sigma}\sqrt{2ln(1.25/\delta)}}{\epsilon}$. Since the values of $\Sigma \in [-1,1]$, then the $l_{2}$-sensitivity of $\Sigma$ is $\triangle_{\Sigma}=2$ in this paper. We give the privacy guarantee of our Fed-TDA as follows.

\textbf{Theorem 2.} \textit{Fed-TDA is $(\epsilon, \delta)$-differential privacy.}
The proof of Theorem 2 follows the standard analysis of Gaussian mechanism in \cite{bassily2019linear}.

\section{Experiments}

\subsection{Experimental Setup}

\subsubsection{Datasets}

We evaluate our Fed-TDA on five real-world tabular datasets, targeting binary and multi-class classification. The list of datasets is as follows: (1) \textbf{Clincial} \cite{chicco2020machine} is a dataset about Cardiovascular diseases (CVDs) that predicts mortality from heart failure information; (2) \textbf{Tuberculosis} \cite{warnat2021swarm} is an RNA-seq dataset based on whole blood transcriptomes, which is used to predict tuberculosis from genetic information; (3) \textbf{CovType} \cite{blackard1999comparative} is for forest cover types prediction; (4) \textbf{Intrusion} is collected from the kddcup99 \cite{tavallaee2009detailed} dataset and contains the top 10 types of network intrusions; (5) \textbf{Body} \cite{kaggle2022} dataset comes from Kaggle and is provided by Korea Sports Promotion Foundation for predicting body performance. Clinical and Tuberculosis datasets are for binary classification, and CovType, Intrusion, and Body datasets are for multi-class classification.

To simulate the non-IID data, we follow the label skew method in \cite{li2022federated}, where each client is allocated a proportion of the samples of each label according to the Dirichlet distribution. Specifically, we allocate a $p_{k,l}$ ($p_{k,l} \sim Dir(\beta)$) proportion of the samples of label $l$ to participant $k$, where $Dir(\beta)$ represents the Dirichlet distribution with a parameter $\beta$ ($\beta > 0$). Then, we can flexibly control the degree of the non-IID by varying the parameter $\beta$. The smaller the $\beta$ is, the higher the degree of label distribution drift. More details on the presentation, statistics, and partitioning of the datasets are given in Appendix {\color{red}E}.

\subsubsection{Baseline Methods}

We compare our Fed-TDA  with five state-of-art methods: one traditional FL method, \textbf{FedAvg} \cite{mcmahan2017communication}; one algorithm-based method, \textbf{FedProx} \cite{li2020federatedopt}; one data sharing-based method, \textbf{Fedmix} \cite{yoon2020fedmix}; and two GAN-based methods, \textbf{DP-FedAvg-GAN} \cite{augenstein2019generative} and \textbf{Fed-TGAN} \cite{zhao2021fed}. Besides FedAvg (\textbf{Fed-TDA + FedAvg}), we also extended our Fed-TDA to Fedprox (\textbf{Fed-TDA+FedProx}). For the classifier, we adopt a simple deep learning model with fully-connected layers for the tabular data. The metric of the binary classification task is ROCAUC, and the multi-class classification task is accuracy. More implementation details are given in Appendix {\color{red}E}.

\subsection{Performance on Non-IID Federated Learning}

\textbf{Varying Non-IID Settings} \quad Table \ref{tab:binary} and Table \ref{tab:multiclass} show the test performance (\textit{mean±std}) of binary and multi-class classification tasks under the different non-IID settings, respectively. In almost all cases, our Fed-TDA achieves the best performance. In addition, the test performance of other methods increases as $\beta$ increases, while Fed-TDA is always stable and close to IID data. This shows that our Fed-TDA can effectively eliminate the non-IID problem by synthesizing high-quality data for data augmentation. For GAN-based methods and Fedmix, most of them perform well in some cases compared with FedAvg. However, there is still a performance gap compared with Fed-TDA because they are limited in the quality of synthetic tabular data under non-IID data. For FedProx, it performs similarly to FedAvg because it does not fundamentally eliminate the distribution shift problem, which is consistent with the conclusion in \cite{li2022federated}. 

\begin{figure}[t]
\centering
\includegraphics[width=\linewidth]{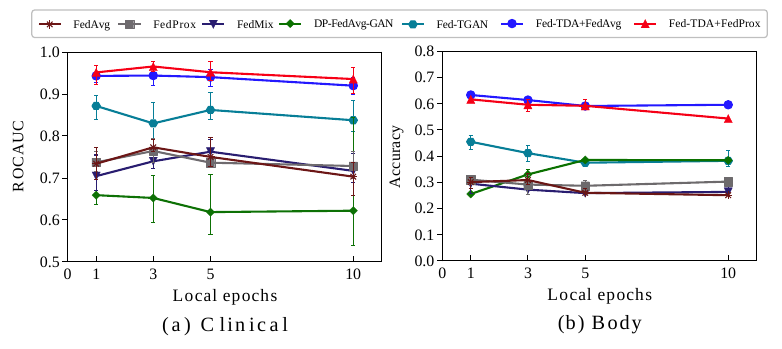}
\caption{Test performance under varying number of local epochs on the Clinical and Body datasets. Where $\beta =0.05$ and $K=5$.} 
\label{fig:epochs} 
\end{figure}

\begin{figure}[t] 
\centering
\includegraphics[width=\linewidth]{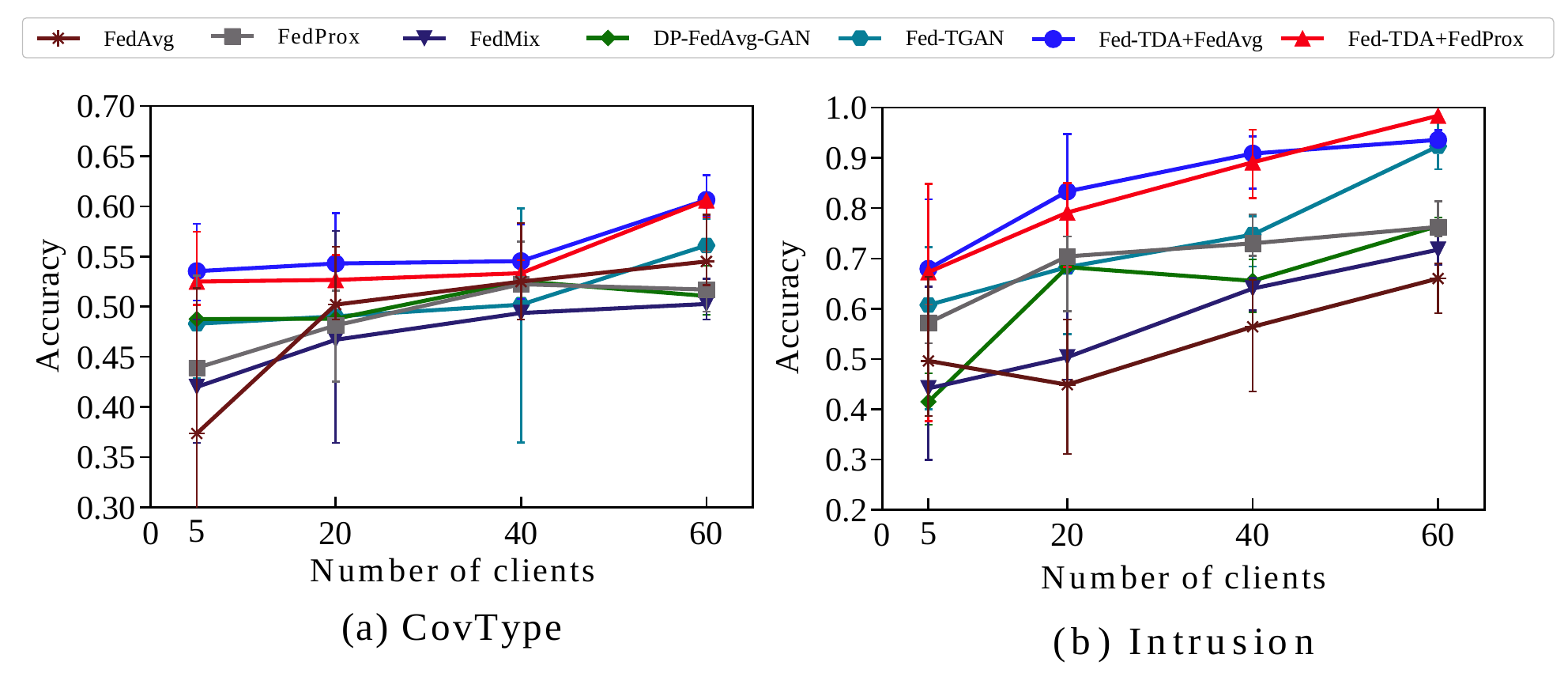}
\caption{Test performance under varying number of clients on the CovType and Intrusion datasets. We set $\beta=0.05$, the local epoch is $3$, and the number of samples per client is kept constant. $K$ varies from 5 to 60. } 
\label{fig:clients} 
\end{figure}

\textbf{Varying Local Epochs} \quad We study the impact of the number of local epochs on test performance. Figure \ref{fig:epochs} shows the test performance (ROCAUC or accuracy) of baseline methods on the Clinical and Body datasets (5 times per case). As the number of local epochs increases from 1 to 10, Fed-TDA achieves the best and most robust performance. It shows the robustness of our Fed-TDA to local training parameters. More results on other datasets are given in Appendix {\color{red}F}.



 









 

\begin{table*}[t]
\centering
\footnotesize

\begin{tabular}{ccccccc|cccccc}
\toprule
 \multirow{3}{*}{\textbf{Methods}} & \multicolumn{6}{c}{\textbf{Clinical} } &  \multicolumn{6}{c}{\textbf{Intrusion} } \cr

 \cmidrule(lr){2-7} 
 \cmidrule(lr){8-13}

 & \multicolumn{2}{c}{ $\beta=0.05$} & \multicolumn{2}{c}{ $\beta=0.1$} & \multicolumn{2}{c}{ $\beta=0.5$} & \multicolumn{2}{c}{ $\beta=0.01$} & \multicolumn{2}{c}{ $\beta=0.05$} & \multicolumn{2}{c}{ $\beta=0.1$}  \cr

\cmidrule(lr){2-3}
\cmidrule(lr){4-5}
\cmidrule(lr){6-7}
\cmidrule(lr){8-9}
\cmidrule(lr){10-11}
 \cmidrule(lr){12-13}

 & JSD & WD & JSD & WD & JSD & WD &JSD & WD &JSD & WD &JSD & WD \cr
 
\midrule
DP-FedAvg-GAN & 0.422&	0.308 & 0.303&	0.317&0.345&	0.318 & 0.588&0.259& 0.510&	0.312&0.618& 0.232 \cr
Fed-TGAN & 0.165&	0.095&0.164&0.102&0.132&0.128& 0.278&	0.279&	0.211&	0.264&	0.181&	0.254 \cr

Fed-TDA & \textbf{0.082}&	\textbf{0.091}&	\textbf{0.081}&	\textbf{0.081}&	\textbf{0.086}&	\textbf{0.095}& \textbf{0.064}&\textbf{0.093} & \textbf{0.064}&\textbf{0.093} &\textbf{0.064}&\textbf{0.093} \cr

\bottomrule
\end{tabular}
\caption{The statistical similarity for DP-FedAvg-GAN, Fed-TGAN, and Fed-TDA. Where $K=5$.}
\label{tab:simi}
\end{table*}

\begin{table*}[t]
\centering
\footnotesize

\begin{tabular}{ccccccccccc}
\toprule

  \multirow{2}{*}{\textbf{Datasets}} & \multicolumn{2}{c}{\textbf{FedAvg/FedProx}} & \multicolumn{2}{c}{\textbf{Fedmix}} & \multicolumn{2}{c}{\textbf{DP-FedAvg-GAN}} & \multicolumn{2}{c}{\textbf{Fed-TGAN}} & \multicolumn{2}{c}{\textbf{Fed-TDA}}  \cr
 
 \cmidrule(lr){2-3} 
    \cmidrule(lr){4-5}
    \cmidrule(lr){6-7}
    \cmidrule(lr){8-9}
    \cmidrule(lr){10-11}
 
 & U & D & U & D & U & D & U & D & U & D\cr
 
\midrule

Clincial & 367.18 &367.18 & \textbf{0.0039} &\textbf{0.0156}&178.45 & 188.82 & 662 & 662 &0.4358 &0.4402 \cr

Tuberculosis & 18052 & 18052 & \textbf{18.21} & \textbf{72.84} & 70971 & 71179.93& 233000 & 233000 &6274.3 &6284.6 \cr

CovType & 436.55 & 436.55 &3.3525 & 13.4103 & 342.51 &353.8 & 1510 & 1510 & \textbf{0.6951} & \textbf{0.7034} \cr

Intrusion& 450.59 & 450.59 &8.0300 & 32.1200 & 291.73 & 302.74 & 2290& 2290 & \textbf{2.4446} &\textbf{2.4665} \cr

Body & 354.15 & 354.15 & \textbf{0.2292} & 0.9168 & 174.54 & 184.89 & 1010 & 1010 & 0.6204 & \textbf{0.6263} \cr

\bottomrule
\end{tabular}
\caption{The total communication costs (MB) of federated learning."U" and "D" are short for "Upload" and "Download", respectively. Where $\beta=0.05$ and $K=5$.}
\label{tab:cost}
\end{table*}

\textbf{Varying Local Clients} \quad Figure \ref{fig:clients}  investigates the impact of the number of clients on test performance on the CovType and Intrusion datasets (5 times per case). Each client kept a fixed number of samples. As the number of clients increases from 5 to 60, the test performance of all methods improves. This is because the amount of data and features will increase with the number of clients. In almost all cases, we observe the superior performance of Fed-TDA compared to other baseline methods. 


\begin{figure}[t] 
\centering
\includegraphics[width=0.95\linewidth]{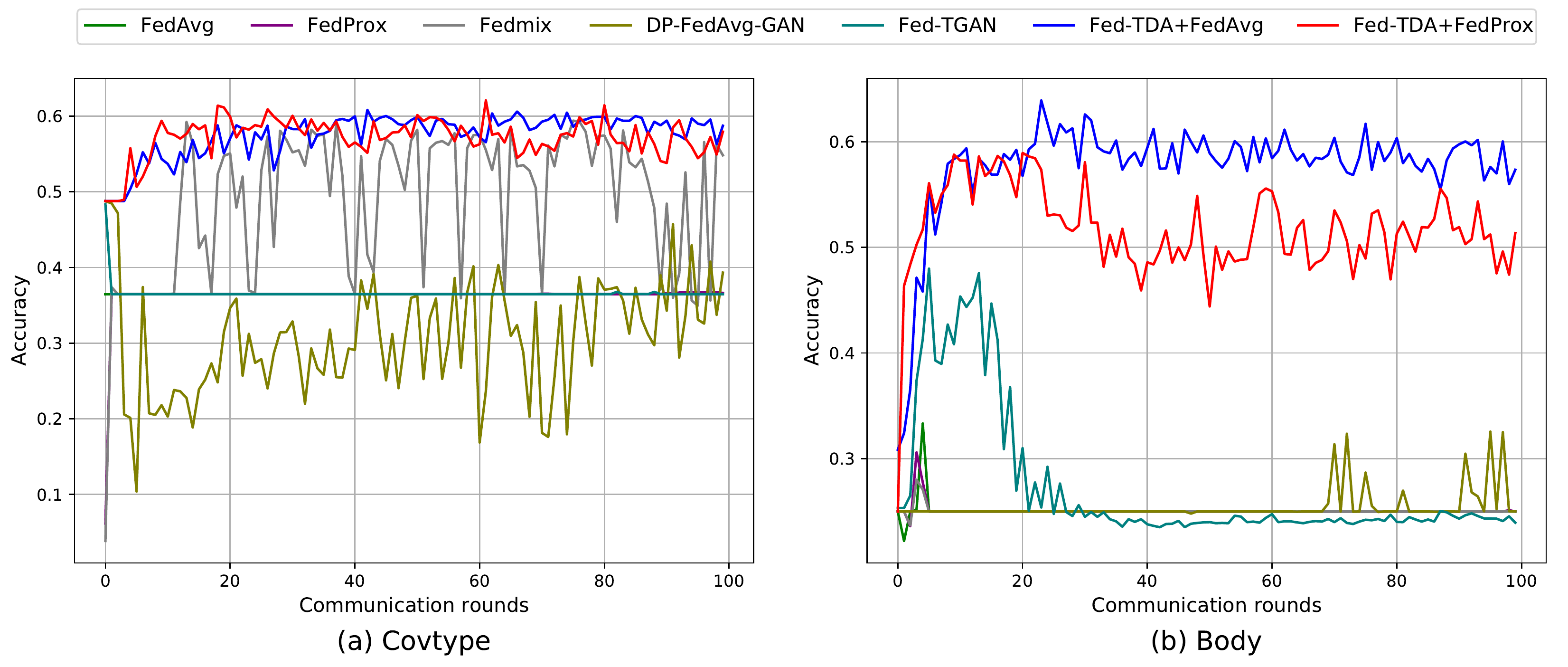}
\caption{Learning curves of the baseline methods on the CovType and Body datasets. Where $\beta=0.05$, $K=5$, the local epoch is $3$.} 
\label{fig:learning} 
\end{figure}

\textbf{Learning Curves} \quad Figure \ref{fig:learning} shows the learning curves on the CovType and Body datasets. As shown in Figure \ref{fig:learning}, Fed-TDA achieves the fastest convergence and the best performance in the non-IID scenario. This is because the proposed Fed-TDA converts non-IID data to IID data by providing high-quality synthetic data. In contrast, the baseline methods do not converge or or converge to a poor state. This indicates that the baseline methods cannot train a usable model.

\subsection{Privacy Study}
\begin{figure}[t] 
\centering
\includegraphics[width=\linewidth]{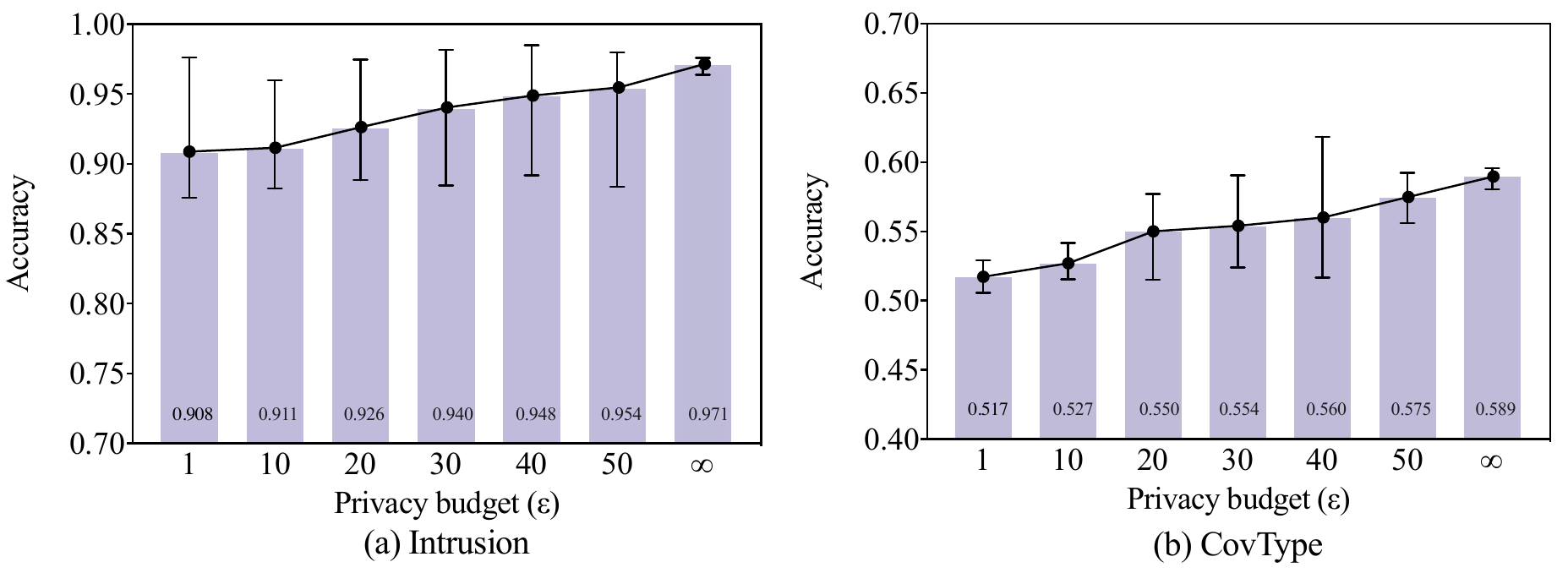}
\caption{Trade-off between the test performance and the privacy level on the Intrusion and CovType datasets. Where $\beta=0.05$, $K=5$, the local epoch is $3$, privacy budget $\delta=10^{-4}$ and $\epsilon$ varies from 1 to $\infty$. } 
\label{fig:privacy} 
\end{figure}

Figure \ref{fig:privacy} demonstrates the trade-off between the test performance of Fed-TDA and the privacy level on the Intrusion and CovType datasets. As shown in Figure \ref{fig:privacy}, the test performance of our Fed-TDA improves as $\epsilon$ varies from 1 to $\infty$. This is because the smaller $\epsilon$ is, the better the privacy but the lower the data utility. As $\epsilon$ increases from $1$ to $\infty$, the test accuracy of Fed-TDA increases from 0.908 to 0.97 on the Intrusion dataset, and from 0.517 to 0.589 on the CovType dataset. More results on other datasets are shown in Appendix {\color{red}F}.

\subsection{Satistical Similarity}

Following \cite{zhao2021ctab}, we use the average Jensen-Shannon Divergence (JSD) and the average Wasserstein Distance (WD) to quantitatively measure the statistical similarity between the synthetic and real data. The average JSD and average WD are used to measure discrete and continuous columns, respectively. Table \ref{tab:simi} shows the statistical similarity results on the Clinical and Intrusion datasets. Compared with GAN-based methods, Fed-TDA consistently achieves the best similarity scores because the synthetic data generated by Fed-TDA keep the same statistical distribution as the real data as discussed in Section \ref{sec:utility}. In addition, as the $\beta$ increases, the statistical similarity scores of GAN-based methods also get better. This shows that GAN-based methods are affected by Non-IID data. Unlike GAN-based methods, our Fed-TDA is a statistical model and will not be affected by non-IID data. More results on other datasets are given in appendix {\color{red}F}.

\subsection{Communication Costs}

Table \ref{tab:cost} shows the total communication costs for $100$ communication rounds of federated training. In practice, the communication rounds of DP-FedAvg-GAN, Fed-TGAN, and Federated VB-GMM are 100, while Fedmix only exchanges once before training. Compared with the algorithm-based method (FedProx) and GAN-based methods (DP-FedAvg-GAN and Fed-TGAN), the communication overhead of Fed-TDA is negligible. This is because they transmit the parameters of models or GANs iteratively, whereas Fed-TDA only exchanges low-dimensional statistics. Compared with Fedmix, Fed-TDA consumes less communication overhead on Covtype, Intrusion, and Body datasets, but more on others. Especially on the Tuberculosis dataset, the communication overhead of Fed-TDA far exceeds that of Fedmix. As analyzed in Appendix {\color{red}D}, the communication costs of Fed-TDA is $O(n_{c}^{2}+n_{d}^{2})$. The Tuberculosis dataset contains 1240 samples but 18136 continuous columns, which makes Fed-TDA consume far more communication overhead than Fedmix. Therefore, a limitation of Fed-TDA is that it will incur a lot of additional communication overhead on high-dimensional data.

\section{Conclusion}

In this paper, we presented Fed-TDA, a federated tabular data augmentation method for solving the non-IID problem in federated learning. Unlike GAN-based methods, Fed-TDA synthesizes tabular data only using some simple statistics, including distributions of each column and global covariance. In Fed-TDA, the MDT and the ICDM are proposed to synthesize continuous columns and discrete columns, respectively. The effectiveness, privacy, and communication efficiency of Fed-TDA have been demonstrated from both theoretical and experimental perspectives.

\bibliography{main}

\newpage

\section{Appendix}

\renewcommand\thesection{\Alph{section}}
\setcounter{secnumdepth}{1} 

\section{Federated VB-GMM}



Gaussian Mixture Model (GMM) \cite{corduneanu2001variational} is a probabilistic model as shown in Equation (\ref{equ:gmm}). 

\begin{equation} 
p(x|\pi, \mu, \Sigma)=\sum_{t=1}^{T}\pi_{t}N(x|\mu_{t},\Sigma_{t})
\label{equ:gmm}
\end{equation}
where $\pi_{t}$ is called mixing coefficients, represents the mixture weight of \textit{t}-th Gaussian distribution. $N(x|\mu_{t},\Sigma_{t})$ represents \textit{t}-th Guassian distribution with mean $\mu_{t}$ and variance $\Sigma_{t}$. GMM is expressed as the sum of the number of $K$ Gaussian distributions multiplied by the corresponding weight $\pi_{t}$. There are four parameters $\pi, \mu, \Sigma$ and $K$ to be optimized when GMM is applied to a dataset.

Variational Bayesian Gaussian Mixture Model (VB-GMM) \cite{corduneanu2001variational} is a model which applying Variational Bayesian method to estimate the parameters of GMM. It introduces three distributions, Dirichlet, Gaussian, and Wishart as the prior distributions for $\pi, \mu$, and $\Lambda$ (inverse matrix of $\Sigma$), as  shown in Equations (\ref{equ:dir}), (\ref{equ:mu}),(\ref{equ:la}) respectively.

\begin{equation} 
\pi \sim Dir(\alpha)
\label{equ:dir}
\end{equation}

\begin{equation} 
\mu \sim N(m,(\beta\Lambda)^{-1})
\label{equ:mu}
\end{equation}

\begin{equation} 
\mathcal{W}(W, \nu)
\label{equ:la}
\end{equation}
where $\alpha$ is a parameter of the Dirichlet distribution, which represents the prior distribution of $\pi$, $m$ and $\beta$ are parameters of the Gaussian distribution, which represent the prior distribution of $\mu$, $W$, and $\nu$ are parameters of the Wishart distribution which represent the prior distribution of $\Lambda$. The parameters of GMM can be calculated by introducing the prior distribution of each parameter combined with the Variational EM algorithm \cite{bishop2006pattern,corduneanu2001variational}.

In this paper, we customize the variational Bayesian GMM algorithm \cite{corduneanu2001variational} under federated learning to learn the global GMM for each continuous column in decentralized tabular data $X=\{x_{1},...,x_{n}\}$. The procedures of federated VB-GMM are as follows:

\textbf{Step 1 (Initialization): Server}
\begin{itemize}
    \item Initialize $\alpha$, $m$, $\beta$, $\Lambda$, $W$, $\nu$, $T$ randomly, where $\alpha$ is a parameter of Dirichlet distribution of which is a prior distribution of $\pi$ ($\pi \sim Dir(\alpha)$), $m$ and $\beta$ are parameters of a Gaussian distribution which is a prior distribution of $\mu$ ($\mu \sim N(m,(\beta \Lambda)^{-1})$), $W$ and $\nu$ are parameters of a Wishart distribution which is a prior distribution of an inverse matrix $\Lambda$ ($\Lambda \sim \mathcal{W}(W, \nu)$).
    \item Send $\alpha, m, \beta, \Lambda, W, \nu, T$ to each Client.
\end{itemize}

\textbf{Step 2 (Variational E step): Client}
 \begin{itemize}
    \item Each client get $\alpha, m, \beta, \Lambda, W, \nu, T$
    \item Update the precision $\Lambda$
        \begin{equation} 
        ln\tilde{\Lambda}=\sum_{d=1}^{D}\psi(\frac{\nu_{t}+1+d}{2})+Dln2+ln|W_{k}|
        \label{equ:lambda}
        \end{equation}
    \item Compute the local responsibilities $r_{nt}$ 
        \begin{equation} 
        ln\tilde{\pi} = \psi(\alpha_{t})-\psi(\sum_{i=1}^{T}(\alpha_{i}))
        \label{equ:pi}
        \end{equation}
       
        \begin{equation}
        \begin{aligned}
            r_{nt}=&\pi_{t}|\Lambda_{t}|^{1/2} \times\\
        & exp\{-\frac{D}{2\beta_{t}}-\frac{\nu_{t}}{2}(x_{n}-m_{t})^{T}W_{t}(x_{n}-m_{t})\}
        \end{aligned}
        \label{equ:rnk}
        \end{equation}
       
    \item Compute the mean $\bar{x}_{t}^{'}$ of local data which belongs to the \textit{t}-th mode
        \begin{equation} 
        \bar{x}_{t}^{'}=\sum_{n=1}^{N_{client}}r_{nt}x_{n}
        \label{equ:localnk}
        \end{equation}
    \item Send $HE(r_{nt})$ and $HE(\bar{x}_{t}^{'})$ to the server  \qquad //$HE(.)$ represents homomorphic encryption.
\end{itemize}

\textbf{Step 3 (Variational E step): Server}
 \begin{itemize}
    \item Get $HE(r_{nt})$ and $HE(\bar{x}_{t}^{'})$ from each client
    \item Compute the global responsibilities $N_{t}$
        \begin{equation} 
        HE(N_{t})= \sum_{c \in C}^{C}HE(r_{nt})
        \label{equ:nk}
        \end{equation}
    \item Update the global mean $\bar{x}_{t}$ of the global data which belongs to the \textit{t}-th mode 
        \begin{equation} 
        HE(\bar{x}_{t})= \frac{1}{N_{t}}\sum_{c \in C}^{C}HE(\bar{x}_{t}^{'})
        \label{equ:xk}
        \end{equation}
    \item Send $HE(N_{t}) $ and $HE(\bar{x}_{t})$ to each client
\end{itemize}
\textbf{Step 4 (Variational M step): Client}
 \begin{itemize}
    \item Each client decrpts $N_{t}$ and $\bar{x}_{t}$
    \item Update $\alpha_{t}, \beta_{t}, \nu_{t}$, $m_{t}$
        \begin{equation} 
        \alpha_{t}= \alpha_{0}+N_{t}
        \label{equ:alphak}
        \end{equation}
        \begin{equation} 
        \beta_{t}= \beta_{0}+N_{t}
        \label{equ:betak}
        \end{equation}
        \begin{equation} 
        \nu_{t}= \nu_{0}+N_{t}
        \label{equ:nuk}
        \end{equation}
        \begin{equation} 
        m_{t}= \frac{1}{\beta_{t}}(\beta_{0}m_{0}+N_{t}\bar{x}_{t})
        \label{equ:nuk}
        \end{equation}
    \item Compute the covariance matrix $S_{t}^{'}$ of the local data which belongs to \textit{t}-th mode
        \begin{equation} 
            S_{t}^{'}= \sum_{n=1}^{N_{c}}r_{nt}(x_{n}-\bar{x}_{t})(x_{n}-\bar{x}_{t})^{T}
        \label{equ:sk1}
        \end{equation}

    \item Send $HE(S_{t}^{'})$ to the server
\end{itemize}

\textbf{Step 5 (Variational M step): Server}
\begin{itemize}
    \item Get $HE(S_{t}^{'})$ from each client
    \item Update the global covariance matrix $S_{t}$ by $S_{t}^{'}$ from each client c $\in C$
        \begin{equation} 
            HE(S_{t})=\frac{1}{N_{t}}\sum_{c \in C}^{C}HE(S_{t}^{'})
        \label{equ:sk}
        \end{equation}
    \item Send $HE(S_{t})$ to each client
\end{itemize}
\textbf{Step 6 (Convergence check): Client}
\begin{itemize}
    \item Each client decrypts $S_{t}$ from the server
    \item Update the parameter $W$
        \begin{equation} 
            W_{t}^{-1}=W_{0}^{-1}+\frac{\beta_{0}N_{t}}{\beta_{0}+N_{t}}(\bar{x}_{t}-m_{0})(\bar{x}_{t}-m_{0})^{T}+N_{t}S_{t}
        \label{equ:wk}
        \end{equation}
    \item Update the lower bound $L$ 
        \begin{equation} 
        \begin{aligned}
            L=&-\sum_{n=1}^{N_{c}}\sum_{t=1}^{T}(e^{r_{nt}} \times r_{nt})\\
                &-\sum_{t=1}^{T}(\frac{\nu_{t}Dln2}{2}-\sum_{t=1}^{T}ln\Gamma(\nu_{t}))\\
                &-(ln\Gamma(\sum_{t=1}^{T}\alpha_{t})-\sum_{t=1}^{T}ln\Gamma(\alpha_{t}))\\
                &-\frac{D\sum_{t=1}^{T}ln\beta_{t}}{2}-\sum_{t=1}^{T}(\nu_{t}|W_{t}|)-\sum_{t=1}^{T}(\nu_{t}|W_{t}|)
        \label{equ:l}
        \end{aligned}
        \end{equation}
    \item Send $L$ to the server
\end{itemize}

\textbf{Step 7 (Convergence check): Server}
\begin{itemize}
   \item Get the lower bound $L$ from the client in this iteration and compare it with the last $L_{last}$. If the difference between them is less than a predefined value $\epsilon$, the process ends. Otherwise, return to the Step 2. 
\end{itemize}

\section{Derivation of Global Mean and Covariance}

We suppose that each client possesses a local tabular dataset $X_{k}=[x_{k,j}]^{N_{k}} \in \mathcal{R}^{N_{k} \times l}$, where $N_{k}>1$ represents the number of samples of $k$-th client. The global dataset is $X=(X_{1},X_{2},...,X_{K})$, where $K$ indicates the number of clients. Then, the global mean is

\begin{equation} 
\begin{split}
E(X) = \frac{1}{N}\sum_{k=1}^{K}\sum_{j=1}^{N_{k}}x_{k,j} 
&=\sum_{k=1}^{K}\frac{N_{k}}{N} \cdot \frac{1}{N_{k}}\sum_{j=1}^{N_{k}}x_{k,j} \\
&=\sum_{k=1}^{K}\frac{N_{k}}{N}E(X_{k}) 
\end{split}
\label{equ:gmean}
\end{equation}
where $E(X) \in \mathcal{R}^{1 \times l}$. The global covariance is 

\begin{equation} 
\begin{split}
\Sigma_{X} &= E[(X-E(X))^{T}(X-E(X))] \\
&=E(X^{T}X)-E^{T}(X)E(X)
\end{split}
\label{equ:gsigma}
\end{equation}

For $X^{T}X$, we have
\begin{equation} 
\begin{split}
X^{T}X &= \begin{bmatrix}
X_{1}^{T}, X_{2}^{T}, \dots ,X_{K}^{T}
\end{bmatrix}
\begin{bmatrix}
X_{1} \\
X_{2} \\
\vdots \\
X_{K}
\end{bmatrix} \\
&= \sum_{k=1}^{K}X_{k}^{T}X_{k}=\sum_{j=1}^{N_{k}}x_{k,j}^{T}x_{k,j} \in \mathcal{R}^{l \times l}
\end{split}
\label{equ:xx}
\end{equation}

Then, we can compute $E(X^{T}X)=E(\sum_{k=1}^{K}X_{k}^{T}X_{k})=\sum_{k=1}^{K}E(X_{k}^{T}X_{k})$. Finally, the global covariance is 

\begin{equation} 
\begin{split}
\Sigma_{X} &= E(X^{T}X)-E^{T}(X)E(X) \\
&=\sum_{k=1}^{K}E(X_{k}^{T}X_{k}) - E^{T}(X)E(X) \\
&= \sum_{k=1}^{K}E(X_{k}^{T}X_{k}) -(\sum_{k=1}^{K}\frac{N_{k}}{N}E(X_{k}))^{T}(\sum_{k=1}^{K}\frac{N_{k}}{N}E(X_{k}))
\end{split}
\label{equ:gsigma1}
\end{equation}

From Equation (\ref{equ:gsigma1}), to obtain the global covariance, we need compute $E(X_{k})$ and $E(X_{k}^{T}X_{k})$ in each client.

\begin{table*}[t]
\centering
\footnotesize

\begin{tabular}{ccccccc}
\toprule
Dataset & \makecell[c]{Number of \\ samples} & \makecell[c]{Number of \\ training sets} & \makecell[c]{Number of \\ test set}& \makecell[c]{Number of \\ continuous columns} & \makecell[c]{Number of \\ discrete columns} & \makecell[c]{ Number of \\ classes} \cr
 
\midrule

Clinical & 299 & 209 & 90 & 7 & 5 & 2 \cr

Tuberculosis & 1550 & 1340 & 310 & 18135 & 0 & 2   \cr

CovType & 581012 & 406704  & 174308 & 10 & 44 & 7  \cr

Intrusion & 29991 & 20000 & 9991 & 38 & 3 & 10 \cr

Body & 13393 & 9373 & 4020 & 10 & 1 & 4 \cr
 
\bottomrule
\end{tabular}
\caption{Statistic of datasets.}
\label{tab:dataset}
\end{table*}

\begin{figure*}[t] 
\centering
\includegraphics[width=\linewidth]{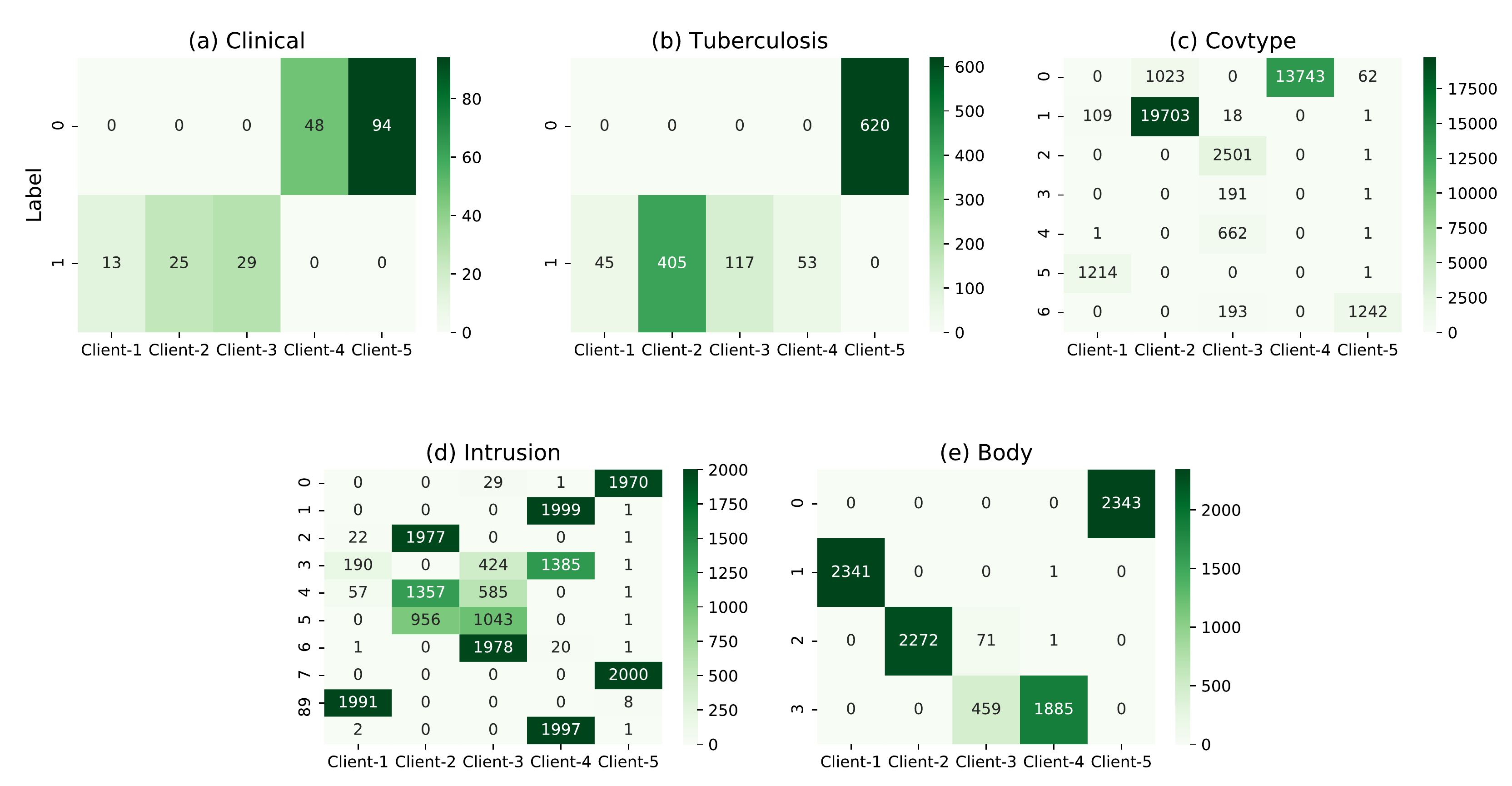}
\caption{Label distributions of Clincial, Tuberculosis, CovType, Intrusion, and Body across the clients, where $\beta=0.05$.} 
\label{fig:data} 
\end{figure*}

\section{Proof of Theorem 1}

\textbf{Theorem 1.} \textit{$X \sim N(0,I), X \in \mathcal{R}^{N \times l}$, the covariance of $X$ is $\Sigma_{X}$ and its Cholesky decomposition is $\Sigma_{X}=UU^{T}$. For $\forall L \sim N(0,I), L \in \mathcal{R}^{N \times l}$. $Y=LU^{T} \in \mathcal{R}^{N \times l}$, we have $\Sigma_{X} = \Sigma_{Y}$ }

\textit{Proof.}
\begin{equation} 
\begin{split}
\Sigma_{Y} &=  E[(Y-E(Y))^{T}(Y-E(Y))]\\
&= E[(Y^{T}-E(Y)^{T})(Y-E(Y))]\\
&= E[Y^{T}Y-E(Y)^{T}Y-Y^{T}E(Y)+E(Y)^{T}E(Y)]\\
&= E[UL^{T}LU^{T}-UE(L)^{T}LU^{T}-UL^{T}E(L)U^{T} \\
&+UE(L)^{T}E(L)U^{T}]\\
&=UE[L^{T}L-E(L)^{T}L-L^{T}E(L)+E(L)^{T}E(L)]U^{T}\\
&=U\Sigma_{L}U^{T}
\end{split}
\label{equ:proofxsy}
\end{equation}
where $\Sigma_{L}$ represents the covariance of $L$. Since that $L$ follows an independent standard normal distribution. Thus $\Sigma_{L} = I$. Then, we have,
\begin{equation} 
\begin{split}
\Sigma_{Y} = U\Sigma_{L}U^{T} =UU^{T} = \Sigma_{X}
\end{split}
\label{equ:proofxsy1}
\end{equation}

\section{Computation complexity and Communication Costs}

Since Fed-TDA introduces MDT, ICDM, and privacy preserving data synthesis on the vanilla FL framework, additional computation and communication overhead are incurred. In this section, we only discuss the additional computation and communication costs introduced by Fed-TDA, compared to vanilla FL. 

\textbf{Computation complexity} \quad On the server-side, as shown in Algorithm 1, the additional computation including federated GMM for each continuous columns, the global frequency for each discrete columns, and global covariance $\Sigma$. The computation complexity of federated GMM on the server is $O(Kmtn_{c})$,  the global frequency is $O(Kn_{d})$, and the global covariance is  $O(Kl^{2})$, where $l=(2n_{c}+n_{d})$. Thus, the total computation complexity of Fed-TDA on the server is $O(K(4n_{c}^{2}+4n_{c}n_{d}+n_{d}^{2}+mtn_{c}+n_{d}))$. On each client, as shown in Algorithm 1 and 2, the additional computation includes local parameters for federated GMM, local frequency for each discrete columns, DMT and ICDM for local data representation, local expectation for global covariance in Equation (\ref{equ:gsigma1}), and privacy preserving data synthesis. The computation complexity for the above process are $O(N_{k}mtn_{c})$, $O(N_{k}n_{d})$, $O(N_{k}l)$, $O(N_{k}l)$, and $O(N_{k}l)$, respectively. Therefore, the total additional computation complexity of each client is $O(N_{k}(6n_{c}+4n_{d}+mtn_{c}))$.

\textbf{Communication costs} \quad As shown in Figure 1, the exchanged statistics include the global covariance, federated VB-GMM, and ICDM. Specifically, for ICDM, each client uploads the local frequency for each category to the server and incurs $n_{d}p$ cost, where $p$ represents the average number of categories for each discrete column. For federated VB-GMM, the communication cost of uploading is $(3t+2)n_{c}m$, where $t$ represents the average number of Gaussian distributions for each continuous column and $m$ is the number of learning rounds of federated VB-GMM. To compute global covariance, each client upload $E(X_{k})$ and $E(X_{k}^{T}X_{k})$ to the server, the cost is $l(l+1)$, where $l=2n_{c}+n_{d}$. Thus, the total communication cost of uploading is $K(4n_{c}^{2}+n_{d}^{2}+4n_{c}n_{d}+(3tm+2m+2)n_{c}+(p+1)n_{d})$. For download, the communication cost of ICDM is $n_{d}p$, the cost for federated VB-GMM is $(6t+1)n_{c}m$, and the cost of global covariance is $l^{2}$. Then, the total communication cost of download is $K(4n_{c}^{2}+n_{d}^{2}+4n_{c}n_{d}+n_{d}p+(6t+1)n_{c}m)$.

With the above analysis, we found that the additional computation and communication cost of Fed-TDA depends on the number of columns (e,g, $n_{c}$ and $n_{d}$). In real-world applications, the number of table columns is much smaller than the number of samples, generally between tens to hundreds. Compared with the federated model training, the additional computation cost and communication overhead of Fed-TDA is almost negligible.

\section{Experimental Details}

\subsection{Datasets}

We evaluate our Fed-TDA on five real-world datasets. Table \ref{tab:dataset} shows the statistics of datasets. The details of these datasets are as follows:

\begin{enumerate}
\item[$\bullet$] \textbf{Clincial} \cite{chicco2020machine} is a dataset about Cardiovascular diseases (CVDs) that predicts mortality from heart failure information, which contains 299 samples with 7 continuous columns and 5 discrete columns. This dataset is used for binary classification task.
\item[$\bullet$] \textbf{Tuberculosis} \cite{warnat2021swarm} is an RNA-seq dataset based on whole blood transcriptomes, which is used to predict tuberculosis from genetic information. This dataset comtains 18135 continuous columns but 1550 samples, which is used for binary classification task.
\item[$\bullet$] \textbf{CovType} \cite{blackard1999comparative} dataset is derived from US Geological Survey (USGS) and US Forest Service (USFS). The label attribute represents forest cover type, which contains seven classes. This dataset contains 581012 samples, where 406704 samples are training set and 174308 are used for testing. We found that the test performance of the model trained by taking 10\% of the training set is close to the whole training data. To facilitate the experiment, we took 10\% of the training set for the federated learning experiment.
\item[$\bullet$]  \textbf{Intrusion} is collected from the kddcup99 \cite{tavallaee2009detailed} dataset and contains the top 10 types of network intrusions. There are 20000 samples for training and 9991 samples for testing.
\item[$\bullet$] \textbf{Body} \cite{kaggle2022} dataset comes from Kaggle and is provided by Korea Sports Promotion Foundation for predicting body performance, which is used for multi-class classification task. This dataset contains 13393 samples with 10 continuous columns and 1 discrete columns, in which 9373 samples are used for training and other 4020 samples are used for testing.
\end{enumerate}

Figure \ref{fig:data} shows the distributions of the above five datasets among the clients. From Figure \ref{fig:data}, we can find that the label distributions are quite heterogeneous. In particular, the number of samples varies among the clients, and each client contains only a few classes of labels.

\subsection{Implementation Details}

\begin{table}[!ht]
\centering
\footnotesize

\begin{tabular}{cl}
\toprule
Layer & Detials \cr
 
\midrule

Input Layer & \makecell[l]{Linear($n_{input}$, 512) \\ BatchNorm1d(512) \\ ReLU() \\ Dropout()}  \cr

\hline

Hidden Layer-1 & \makecell[l]{Linear($512, 256$) \\ BatchNorm1d(256)\\ ReLU() \\ Dropout()}\cr
\hline

Hidden Layer-2 & \makecell[l]{Linear($256, 128$) \\ BatchNorm1d(128)\\ ReLU() \\ Dropout()}\cr

\hline
Hidden Layer-3 & \makecell[l]{Linear($128, 64$) \\ BatchNorm1d(64)\\ ReLU() \\ Dropout()}\cr

\hline

Output Layer & Linear(64, $n_{output}$) \cr
 
\bottomrule
\end{tabular}
\caption{Detailed information of our simple deep learning model. $n_{input}$ represents the dimensional of input data and $n_{output}$ represents the output dimension.}
\label{tab:model}
\end{table}


\begin{table}[!ht]
\centering
\footnotesize

\begin{tabular}{cl}
\toprule
Hyperparameters & Detials \cr
 
\midrule

Communication rounds & 100 \cr

Optimizer & Adam \cr
Learning rate & 0.001 \cr
Weight decay & 1e-5\cr
Local epoch & 3\cr
Batch size & 64\cr
Dropout& 5\cr
Clients per round & 5\cr
$\mu$ for FedProx & 0.05\cr
$M_{k}$ for Fedmix & 5\cr
$\lambda$ for FedProx & 0.05\cr
\bottomrule
\end{tabular}
\caption{Hyperparameters used in our experiments.}
\label{tab:hyperp}
\end{table}

Table \ref{tab:model} shows the details of our simple deep learning model used for classifier. Note that, the goal of our experiment is not to find an optimal classifier. Thus, this vanilla classifier is used for all experiments in this paper. Table \ref{tab:hyperp} shows the detailed hyperparameters used in our experiments.

\section{Extra Experimental Results}

\begin{figure*}[t] 
\centering
\includegraphics[width=\linewidth]{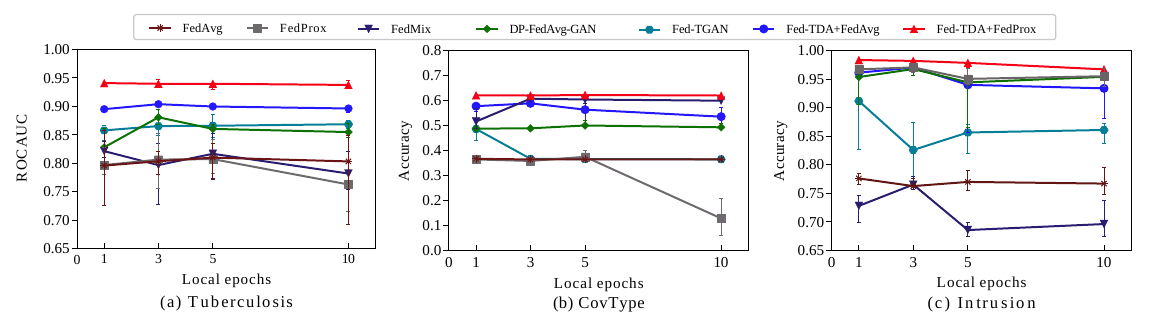}
\caption{Test performance under varying number of local epochs on the Tuberculosis, CovType, and Body datasets. Where $\beta =0.05$ and $K=5$.} 
\label{fig:epoch} 
\end{figure*}

\begin{figure*}[t] 
\centering
\includegraphics[width=\linewidth]{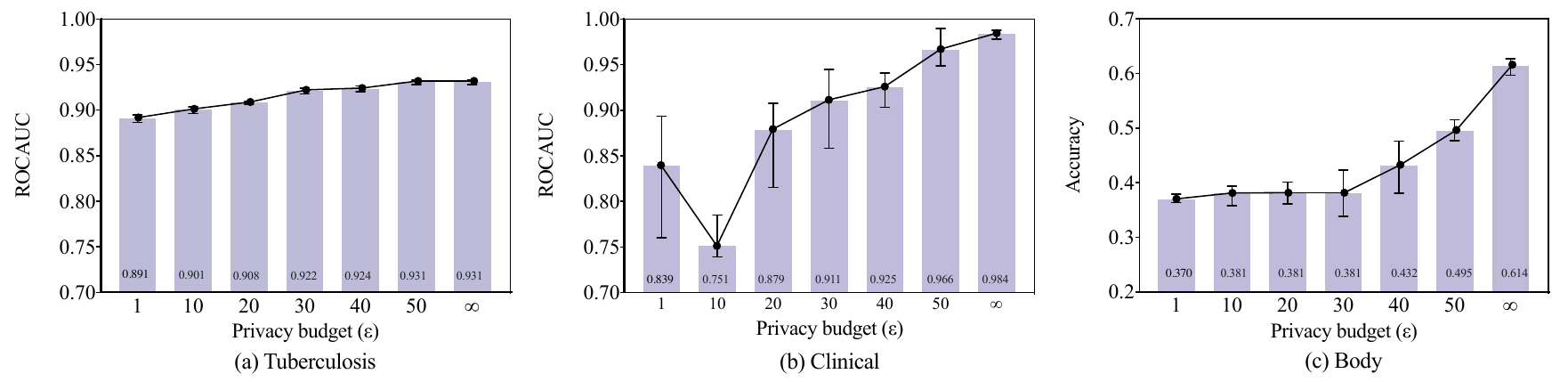}
\caption{Trade-off between the test performance and the privacy level on the Tuberculosis, Clincial, and Body datasets. Where $\beta=0.05$, $K=5$, the local epoch is $3$, privacy budget $\delta=10^{-4}$ and $\epsilon$ varies from 1 to $\infty$.} 
\label{fig:privacy} 
\end{figure*}


\subsection{Varying local epochs}

Figure \ref{fig:epoch} investigates the impact of the number of local epochs on test performance, which shows the average test ROCAUC or accuracy of baseline methods on the Tuberculosis, CovType, and Body datasets. As the number of local epochs increases from 1 to 10, Fed-TDA achieves the best and most robust performance. The experimental results in these datasets are consistent with the other two datasets.



\subsection{Privacy study}

Figure \ref{fig:privacy} demonstrates the trade-off between the test performance of Fed-TDA and the privacy level on the Tuberculosis, Clincial, and Body datasets. The test performance of our Fed-TDA improves as $\epsilon$ varies from 1 to $\infty$.

\subsection{Statistical Similarity}

\begin{table*}[t]
\centering
\footnotesize

  \resizebox{\textwidth}{15mm}{
\begin{tabular}{cccc|cccccc|cccccc}
\toprule
 \multirow{3}{*}{\textbf{Methods}} & \multicolumn{3}{c}{\textbf{Tuberculosis} } &  \multicolumn{6}{c}{\textbf{CovType} }  &  \multicolumn{6}{c}{\textbf{Body} } \cr

 \cmidrule(lr){2-4} 
 \cmidrule(lr){5-10}
\cmidrule(lr){11-16}

 & $\beta=0.05$ & $\beta=0.1$ & $\beta=0.5$ & \multicolumn{2}{c}{ $\beta=0.01$} & \multicolumn{2}{c}{ $\beta=0.05$} & \multicolumn{2}{c}{ $\beta=0.1$} & \multicolumn{2}{c}{ $\beta=0.01$} & \multicolumn{2}{c}{ $\beta=0.05$} & \multicolumn{2}{c}{ $\beta=0.1$}\cr

\cmidrule(lr){2-2}
\cmidrule(lr){3-3}
\cmidrule(lr){4-4}
\cmidrule(lr){5-6}
\cmidrule(lr){7-8}
 \cmidrule(lr){9-10}
\cmidrule(lr){11-12}
\cmidrule(lr){13-14}
 \cmidrule(lr){15-16}

 &  WD & WD &  WD &JSD & WD &JSD & WD &JSD & WD &JSD & WD &JSD & WD &JSD & WD\cr
 
\midrule
DP-FedAvg-GAN & 0.179&	0.192&	0.201 & 0.374&	0.429&0.333&	0.353&0.397&0.260 & 0.134&0.315&	0.150&	0.371&	0.230&0.319\cr

Fed-TGAN &0.178&	0.201&	0.205& \textbf{0.100}&\textbf{0.085}&\textbf{0.099}&0.103&0.110&	0.102 &0.077&	0.080&	0.067&0.075&\textbf{0.037}&	0.076 \cr

Fed-TDA & \textbf{0.141}&	\textbf{0.132}&	\textbf{0.145} & 0.102 & 0.099& 0.102 & \textbf{0.099}&\textbf{0.102} & \textbf{0.099} & \textbf{0.066}&	\textbf{0.068} & \textbf{0.066}&	\textbf{0.068}&0.066&	\textbf{0.068}\cr

\bottomrule
\end{tabular}}
\caption{The statistical similarity for DP-FedAvg-GAN, Fed-TGAN, and Fed-TDA. Where $K=5$.}
\label{tab:simi}
\end{table*}

Table \ref{tab:simi} shows the statistical similarity results of two GAN-based methods on Tuberculosis, CovType, and Body datasets. Since there is only one discrete column in Tuberculosis tables, we compute its average WD. We can find that our Fed-TDA consistently achieves the best similarity scores.

\subsection{Visualisation of Synthesized Samples}

\begin{table*}[!ht]
\centering

\begin{tabular}{ccccccccc}

\toprule
& age & anaemia & phosphokinase & platelets & sex &  smoking & time & label \cr
\midrule
1 & 73 & 0 & 582 & 203000 & 1 & 0 & 195 & 0\cr
2 & 75 & 1 & 246 & 127000 & 1  & 0 & 10 & 1\cr
3 & 50 & 0 & 369 & 252000 & 1 &  0 & 90 & 0\cr
4 & 57 & 1 & 129 & 395000 & 0  & 0 & 42 & 1\cr
5 & 55 & 0 & 109 & 254000 & 1  & 1 & 60 & 0\cr
6 & 45 & 0 & 582 & 166000 & 1  & 0 & 14 & 1\cr
\bottomrule
\end{tabular}
\caption{Sample records in the origin Clinical table of each client.}
\label{tab:origin}
\end{table*}

\begin{table*}[!ht]
\centering

\begin{tabular}{ccccccccc}

\toprule
& age & anaemia & phosphokinase & platelets & sex &  smoking & time & label \cr
\midrule
1 & 51 & 0 & 2120 & 313120 & 1 & 1 & 209 & 0\cr
2 & 51 & 0 & 1425 & 216723 & 0  & 1 & 66 & 1\cr
3 & 59 & 1 & 470 & 272237 & 0 &  0 & 77 & 0\cr
4 & 72 & 1 & 23 & 264638 & 1  & 0 & 59 & 1\cr
5 & 74 & 1 & 309 & 298268 & 0  & 1 & 55 & 0\cr
6 & 56 & 1 & 23 & 368381 & 1  & 0 & 6 & 1\cr
\bottomrule
\end{tabular}

\caption{Sample records in the synthesized Clinical table by Fed-TDA.}
\label{tab:fake}
\end{table*}

We show some synthesized samples in Table \ref{tab:fake}, based on the Clinical dataset, which generated by our Fed-TDA. Some records of the origin Clinical dataset from three clients are shown in Table \ref{tab:origin} after choosing a subset of columns for space consideration. In Table \ref{tab:origin}, rows 1-2 are from Client-1, rows 3-4 from Client-2, and the last two rows are from Client-3. As we can find, there is no one-to-one relationship between Table \ref{tab:origin} and Table \ref{tab:fake}, and the real records have very different values from the synthesized record. It is almost impossible to re-identify the original information from the synthesized data.

\end{document}